\documentclass[MIRU,submit,english]{miru2023e}
\usepackage[]{graphicx}
\usepackage{amsmath}
\usepackage{subfigure}
\usepackage{appendix}
\newtheorem{theorem}{Theorem}[section]
\newtheorem{proposition}[theorem]{Proposition}
\newtheorem{lemma}[theorem]{Lemma}

\newtheorem{definition}[theorem]{Definition}

\begin{document}

\title{New metrics for analyzing continual learners}

\affiliate{Paris1}{Univ Gustave Eiffel}
\affiliate{Paris2}{ESIEE Paris}
\affiliate{Paris3}{LIGM}
\affiliate{Tokyo}{The University of Tokyo}

 \author{Nicolas MICHEL}{Paris1, Paris2, Paris3}[nicolas.michel@esiee.fr]
 \author{Giovanni CHIERCHIA}{Paris1, Paris2, Paris3}[giovanni.chierchia@esiee.fr]
 \author{Romain NEGREL}{Paris1, Paris2, Paris3}[romain.negrel@esiee.fr]
 \author{Jean-François BERCHER}{Paris1,Paris2,Paris3}[jf.bercher@esiee.fr]
 \author{Toshihiko YAMASAKI}{Tokyo}[yamasaki@cvm.t.u-tokyo.ac.jp]

\maketitle

\section*{Abstract}

Deep neural networks have shown remarkable performance when trained on independent and identically distributed data from a fixed set of classes. However, in real-world scenarios, it can be desirable to train models on a continuous stream of data where multiple classification tasks are presented sequentially. This scenario, known as Continual Learning (CL) poses challenges to standard learning algorithms which struggle to maintain knowledge of old tasks while learning new ones. This stability-plasticity dilemma remains central to CL and multiple metrics have been proposed to adequately measure stability and plasticity separately. However, none considers the increasing difficulty of the classification task, which inherently results in performance loss for any model. In that sense, we analyze some limitations of current metrics and identify the presence of setup-induced forgetting. Therefore, we propose new metrics that account for the task's increasing difficulty. Through experiments on benchmark datasets, we demonstrate that our proposed metrics can provide new insights into the stability-plasticity trade-off achieved by models in the continual learning environment.

\section{Introduction}

Although deep neural networks can exhibit remarkable performances when trained on independent and identically distributed data drawn from a fixed set of classes, the practical need arises to train models on a continuous stream of data. In such real-world scenarios, multiple classification tasks are presented sequentially, leading to a situation where the data pertaining to previous tasks becomes inaccessible when learning new ones. This scenario is known as \textit{Continual Learning} (CL) and poses many challenges to standard learning algorithms which can suffer from Catastrophic Forgetting (CF).
Learning algorithms must trade-off between maintaining performances on old tasks (\textit{stability}) and achieving competitive performances on the current task (\textit{plasticity}).
While the main objective remains to maximize accuracy across all classes at the end of training, it is also essential to define meaningful metrics to capture individual methods learning behavior and capabilities.

Metrics such as the Average Accuracy (AA) and the Average Forgetting (AF) have been proposed in past studies. However, none of these metrics takes into account the increasing difficulty of the classification task, which automatically induces a loss in performance for any model. In that sense, we argue that the AF metric is inherently linked to the continual learning setup and does not fairly represent how the model reacts to the continuous environment.

In this paper, we analyze the limitations of the current forgetting metric through simple examples and propose new metrics for CL that take into account the increasing difficulty of the task being solved by the model. We show through several experiments on benchmark datasets that our proposed metrics can shed new light on the \textit{stability-plasticity} trade-off reached by the model when training on the continual environment.
In that sense, we make the following contributions:
\begin{itemize}
    \item We review traditional metrics and show their current limitations;
    \item We propose new metrics which take into account the increasing difficulty of the continual setup;
    \item We experimentally demonstrate the advantages of our metrics compared to traditional metrics for analyzing continual learners.
\end{itemize}
The rest of the paper is organized as follows. In section \ref{sec:related} we describe work related to ours. In section \ref{sec:metric_limitation}, we review two classical CL metrics and show their limitations. In section \ref{sec:new_metrics} we defined our proposed metrics. Section \ref{sec:exp} presents our experiments and eventually section \ref{sec:conclusion} concludes the paper.

\section{Related Work}
\label{sec:related}
\subsection{Online Continual Learning (OCL)}
OCL addresses the problem of learning from a continuous stream of data. Formally, we consider a sequential learning setup with a sequence $\{\mathcal{T}_1,\cdots,\mathcal{T}_K\}$ of $K$ tasks, and $\mathcal{D}_k=(X_k, Y_k)$ the corresponding data-label pairs. In CL, we often assume that for any value $k_1,k_2 \in \{1,\cdots,K\}$, if $k_1\neq k_2$ then we have $Y_{k_1}\cap Y_{k_2}=\emptyset$ and the number of classes in each task is the same. Contrary to standard CL, in OCL only one pass over the data is allowed. This setup has been widely studied in the supervised scenario \cite{aljundi_gradient_2019, aljundi_online_2019, mai_online_2021, rolnick_experience_2019, mai_supervised_2021,vedaldi_gdumb_2020, guo_online_2022}. It is also referred to as Online Class Incremental Learning. This setup is the one considered for experiments in this paper.

\subsection{Memory Based Method}
Memory-based methods consist in using a memory buffer to store a small portion of past samples. When encountering a new batch coming from the stream, another batch is retrieved from the memory, and the model is trained on the combination of both stream and memory batches. Between one stream batch and the other, the memory is updated using the current stream batch data. Memory-based methods have been especially popular in Online Continual Learning as they achieve the best performances \cite{vedaldi_gdumb_2020, mai_online_2021, rolnick_experience_2019, lin_pcr_2023, michel_contrastive_2022, guo_ocm_2022, jung_new_2023}. Compared methods in the experiments section are all memory-based.

\section{Traditional Metrics}
In continual learning for image classification, we are interested in the accuracy of the held-out test sets of the learned classes, just as in standard learning. However, as the model is trained on a sequence of tasks $\{\mathcal{T}_1,\cdots,\mathcal{T}_k\}$, specific metrics have been designed. In this section, we first define traditional metrics used in CL for measuring plasticity and stability. Second, we study their current limitations.

\subsection{Traditional Metrics Definitions}
\begin{definition}
\label{def:aa}
The \textbf{Average Accuracy} (AA) after training on $\mathcal{T}_k$, for a classifier $g$ in the set $\mathcal{G}$ of all classifiers, is:
\begin{equation}
\label{eq:avg_acc}
        AA_k(g) = \frac{1}{k}\sum_{j=1}^k{a_{k,j}(g)},
\end{equation}
with $a_{k,j}(g)$, the accuracy of classifier $g$ on task $j$ after training on $\{\mathcal{T}_1,\cdots,\mathcal{T}_k\}$.
\end{definition}
The final average accuracy $AA_K(g)$ is the accuracy after training $g$ on the last task $\mathcal{T}_K$ and is the metric of interest for evaluating the performance of $g$.

\begin{definition}
\label{def:af} 
    The \textbf{Average Forgetting} (AF) after training on $\mathcal{T}_k$, for a classifier $g \in \mathcal{G}$ is:
\begin{equation}
\label{eq:avg_fgt}
        AF_k(g) = \frac{1}{k-1}\sum_{j=1}^{k-1}{f_{k,j}(g)},
\end{equation}
with $f_{k,j}(g) = \max\limits_{l\in\{j,\cdots,k-1\}}{a_{l,j}(g)} - a_{k,j}(g)$ and $k \ge 2$.
\vskip -0.1in
\end{definition}
Other CL metrics also exist but are not covered in this work. Please note that none of existing metrics take into account task difficulty.

\subsection{Traditional Metrics Limitations}
\label{sec:metric_limitation}
Here, we point out some current limitations of AA and AF by using the example of a random classifier and a toy example.

\textbf{Random classifier case.} Looking at Definition \ref{def:af}, it can be observed that AF refers to how poorly the model currently performs when compared to its all-time-high performance on previous tasks. The main limitation of AF is that it does not solely measure how much the model forgot, but also how much harder the current overall problem is compared to a single task. To illustrate this phenomenon, let us consider a simple example where we learn from a sequence of 5 tasks $\{\mathcal{T}_1,\cdots,\mathcal{T}_5\}$, each task being composed of two classes for a total of 10 distinct classes. We want to use a random classifier $Rand_{C_k}$, where $C_k$ is the total number of classes when training on task $k$.
For this example, the accuracy and forgetting are displayed in Tables~\ref{tab:random_acc} and \ref{tab:random_fgt} respectively.
\begin{table}[t]
    \centering
    \resizebox{0.4\textwidth}{!}{
        \begin{tabular}{c | c c c c c | c}
            $a_{k,j}$      & $\mathcal{T}_1$  & $\mathcal{T}_2$ & $\mathcal{T}_3$ & $\mathcal{T}_4$ & $\mathcal{T}_5$ & $AA_k$ \\ [2pt]
            \hline
            $\mathcal{T}_1$ & $50$ & - & - & - & - & $50$ \\ [2pt]
            $\mathcal{T}_2$ & $25$ & $25$ & - & - & - & $25$ \\ [2pt]
            $\mathcal{T}_3$ & $16.7$ & $16.7$ & $16.7$   & - & - & $16.7$ \\ [2pt]
            $\mathcal{T}_4$ & $12.5$ & $12.5$ & $12.5$ & $12.5$ & - & $12.5$ \\ [2pt]
            $\mathcal{T}_5$ & $10$ & $10$ & $10$ & $10$ & $10$ & $10$
        \end{tabular}
    }
    \caption{\label{tab:random_acc} Accuracy (\%) of a random classifier in a continual setting of 5 tasks with 2 classes per task. Each element at row $k$ and column $j$ is the accuracy $a_{k,j}(Rand_{2k})$}
\end{table}

\begin{table}[t]
    \centering
    \resizebox{0.4\textwidth}{!}{
        \begin{tabular}{c | c c c c c | c}
            $f_{k,j}$      &$\mathcal{T}_1$  & $\mathcal{T}_2$ & $\mathcal{T}_3$ & $\mathcal{T}_4$ &$\mathcal{T}_5$ & $AF_k$ \\ [2pt]
            \hline
            $\mathcal{T}_1$ &- & - & - & - & - & - \\ [2pt]
            $\mathcal{T}_2$ &$25$ & - & - & - & - & $25$ \\ [2pt]
            $\mathcal{T}_3$ &$33.3$ & $8.33$ & - & - & - & $20.83$ \\ [2pt]
            $\mathcal{T}_4$ &$37.5$ & $12.5$ & $4.16$ & - & - & $18.06$ \\ [2pt]
            $\mathcal{T}_5$ &$40$ & $15$ & $6.66$ & $2.5$ & - & $16.04$
        \end{tabular}
    }
    \caption{Forgetting (\%) of a random classifier in a continual setting of 5 tasks with 2 classes per task. Each element at row $k$ and column $j$ is the forgetting $f_{k,j}(Rand_{2k})$}
    \label{tab:random_fgt}
\end{table}
As expected, the accuracy decreases during training as the number of classes becomes larger, therefore the AF is positive. However, a random classifier can hardly forget knowledge as it does not learn at all. While expanding the number of classes should make the task harder for the model, the drop in performance is not due to the model forgetting knowledge but rather to the fact that the model is not able to learn how to solve the hard task as well as the easy one. Using AA and AF can demonstrate forgetting for any model, even if the model cannot possibly forget.

\begin{figure}[t]
\centerline{\includegraphics[width=0.9\columnwidth]{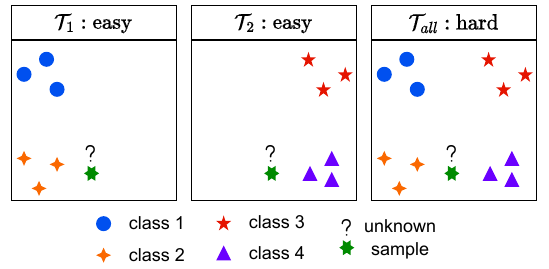}}
\caption{Toy example of a two tasks learning sequence. The model learns from $\mathcal{T}_1$ then $\mathcal{T}_2$ and is then evaluated on $\mathcal{T}_{all}$. Left tile represents learning $\mathcal{T}_1$, middle tile represents learning $\mathcal{T}_2$ and right tile represents learning $\mathcal{T}_{all}=\{\mathcal{T}_1,\mathcal{T}_2\}$. Best viewed in color.}
\label{fig:fgt_limit}
\end{figure}

\textbf{Toy example}. Previous example demonstrates a setup-induced forgetting due to the increasing task difficulty. This increasing task difficulty effect is also illustrated in Figure~\ref{fig:fgt_limit} with a simple 2 tasks sequence example. In this case, the classification tasks $\mathcal{T}_1$ and $\mathcal{T}_2$ are simple. However, the overall task $\mathcal{T}_{all}=\{\mathcal{T}_1,\mathcal{T}_2\}$ is more complex and it is the task the model is evaluated on after training on $\mathcal{T}_2$. It is unlikely that the model forgot how to solve both binary classification tasks separately, as they are very similar. Specifically, in this example, the model might not forget how to differentiate classes 1 and 2 or how to differentiate classes~3 and~4. Rather, the model might not learn how to differentiate classes~1 and~3 or how to differentiate classes~2 and~4.

\section{New Metrics for Continual Learning}
\label{sec:new_metrics}
Current accuracy and forgetting metrics depend strongly on the continual learning setup and can mislead forgetting analysis. We propose two metrics that attempt to dissociate this setup-induced forgetting from the overall performance by rescaling the original AA and AF using the performances of a random classifier to account for task difficulty.
\subsection{Rescaled Average Accuracy and Forgetting}
\begin{definition}
\label{def:uraa1}
    The unnormalized Rescaled Average Accuracy (uRAA) and unnormalized Rescaled Average Forgetting (uRAF) after training on $\mathcal{T}_k$, for a classifier $g$ are:
\begin{equation}
\begin{aligned}
    uRAA_k(g) &= \frac{AA_k(g)}{AA_k(Rand_{C_k})}, \\
    uRAF_k(g) &= \frac{AF_k(g)}{AF_k(Rand_{C_k})}, 
\end{aligned}
\end{equation}
where $C_k$ is the total number of classes seen at task $\mathcal{T}_k$.
\end{definition}

\begin{definition}
The \textit{Rescaled Average Accuracy (RAA)} and \textit{Rescaled Average Forgetting (RAF)} after training on $\mathcal{T}_k$, for a classifier $g \in \mathcal{G}$ are:
\label{def:raa1}
\begin{equation}
\label{eq:raa1}
\begin{aligned}
    RAA_k(g) &= \frac{uRAA_k(g)}{\max\limits_{f\in \mathcal{G},k}{uRAA_k(f)}}, \\
    RAF_k(g) &= \frac{uRAF_k(g)}{\max\limits_{f\in \mathcal{G},k}{uRAF_k(f)}}.
\end{aligned}
\end{equation}
%
\end{definition}

\begin{proposition}
\label{prop:raa}
Let $K$ be the total number of tasks and $C_K$ the total number of classes. The Rescaled Average Accuracy (RAA) after training on $\mathcal{T}_k$, for a classifier $g$ can be expressed as:
\begin{equation}
\label{eq:raa2}
\begin{aligned}
    RAA_k(g)  = \gamma_k AA_k(g)\ \ \  \text{with}\ \ \  \gamma_k =\frac{C_k}{C_K}.
\end{aligned}
\end{equation}
\end{proposition}


\begin{proposition}
\label{prop:raf}
If every task has the same number of classes, the Rescaled Average Forgetting (RAF) after training on $\mathcal{T}_k$, for a classifier $g$ can be expressed as:
\begin{equation}
\begin{aligned}
    &RAF_k(g) = \beta_k AF_k(g), \\
    \text{with}\ \ \ \ \beta_k& = \frac{(H_K-1)(k-1)}{(H_k-1)(K-1)}\ \ \text{and}\ \  H_k = \sum_{i=1}^k \frac{1}{i}.
\end{aligned}
\end{equation}
\end{proposition}
 
\begin{figure}[t]
    \centering
    \subfigure[]{\includegraphics[width=0.45\linewidth]{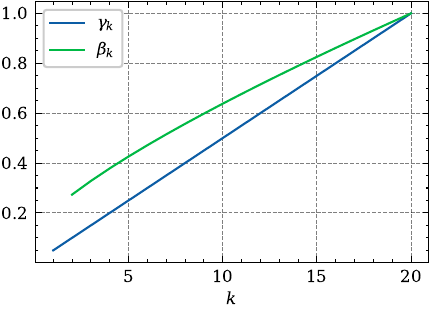}}
    \subfigure[]{\includegraphics[width=0.45\linewidth]{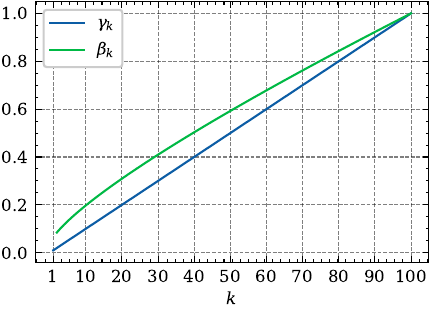}}
    \caption{Visualization of accuracy coefficient $\gamma_k$ and  forgetting coefficient $\beta_k$ for (a) $K=20$   (b) $K=100$}
    \label{fig:beta_gamma}
\end{figure}

The coefficients $\beta_k$ and $\gamma_k$ act as task difficulty coefficients. Figure~\ref{fig:beta_gamma} shows different values for these coefficients along training. The harder the task, the higher the coefficient. Indeed, $(\beta_k,\gamma_k) \in\  \rbrack0,1\rbrack^2$ with one being the maximum difficulty corresponding to the final task. With such definitions, we have equal values at the end of training such that $RAA_K(g)=AA_K(g)$ and $RAF_K(g)=AF_K(g)$.

\subsection{Interpreting RAA and RAF}
\label{sec:interpretation}
In subsequent we give insight on how to interpret previously refined metrics.

\textbf{Comparison to a random classifier.} RAA and RAF represent how much the model learns or forgets when compared to a random classifier.
For example, a constant RAA indicates that the model performances are decreasing exactly like a random classifier. This situation can happen if the model cannot learn new tasks efficiently while maintaining high performance on past tasks (high stability and low plasticity) or oppositely the model perfectly learns new tasks while forgetting older tasks (low stability and high plasticity). In other words, a constant RAA translates a failed stability-plasticity trade-off, while an increasing RAA demonstrates that the model learns more than a random classifier and hence can still accumulate new knowledge despite the increasing task difficulty. Both previous situations cannot be easily differentiated using AA as its value would be decreasing in both cases.

\textbf{Decoupling task difficulty.} As explained in section~\ref{sec:metric_limitation}, the increase in task difficulty while training often implies a decrease in model accuracy. Current RAA and RAF expression can be interpreted as the correction of current AA and AF to compensate for the drop in performance due to task difficulty. Decoupling the task difficulty allows us to detect if the model's performances change due to the increasing number of classes or due to a failed stability-plasticity trade-off.

To showcase how RAA and RAF help detect learning regimes hardly visible with AA and AF, we conduct several experiments, detailed in section \ref{sec:exp}.

\section{Experiments}
\label{sec:exp}
In the following, we apply our metric to several Online Continual Learning methods and compare their values to standard metrics.

\subsection{Datasets}
We use variations of standard image classification datasets \cite{Krizhevsky2009LearningML,le_tiny_2015} to build continual learning environments. The original datasets are split into several tasks of non-overlapping classes. Specifically, we experimented on split-CIFAR100 and split-Tiny ImageNet. In this paper, we omitted the split- suffix for simplicity. \textbf{CIFAR100} contains 50,000 32x32 train images as well as 10,000 test images and is split into 10 tasks containing 10 classes each for a total of 100 distinct classes. \textbf{Tiny ImageNet} is a subset of the ILSVRC-
2012 classification dataset and contains 100,000 $64 \times64$ train images as well as 10,000 test images and is split into 100 tasks containing two classes each for a total of 200 distinct classes.

\subsection{Baselines}
In the following, we describe considered baselines. For every memory-based method, we use reservoir sampling \cite{vitter_random_1985} for memory update and random retrieval. \\
\textbf{Experience Replay} (ER) \cite{rolnick_experience_2019}: ER is a supervised memory based technique using reservoir sampling \cite{vitter_random_1985} for memory update and random retrieval. The model is trained using cross-entropy.\\
\textbf{Finetune}: Usual model training with no measure taken to mitigate forgetting. \\
\textbf{GDumb} \cite{vedaldi_gdumb_2020}: Simple method that stores data from the stream in memory, with the constraint of having a balanced class selection. At inference time, the model is trained offline on memory data.\\

\subsection{Results}
\begin{figure}[t]
    \centering
    \includegraphics[width=1.0\linewidth]{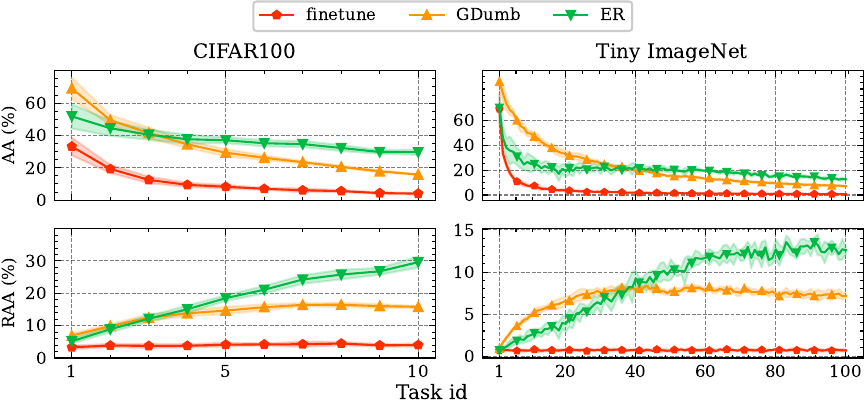}
    \caption{Values of $AA_k$ and $RAA_k$ with $k \in [1, K]$ the task id. From left to write, the performances are shown on CIFAR100 $M=2k$ with $K=5$ and Tiny ImageNet $M=2k$ with $K=5$. The top row shows the AA while the bottom row shows the RAA. Comparing AA to RAA, we can observe different training behavior, notably, GDumb reaches a plateau in CIFAR100 using RAA which is not the case using AA.}
    \label{fig:aa_all}
\end{figure}

In the following, we analyze our experimental results.
\textbf{Training regimes with RAA.} Comparison of obtained values with AA and RAA are displayed in Figure~\ref{fig:aa_all}. It can be observed that for all methods the AA decreases during training, which is expected. Given AA alone, it can be hard to understand what dissociates each method's behavior, as they all follow a similar trend. However, looking at RAA, distinct training regimes can be observed. Notably, GDumb RAA reaches a plateau on Tiny ImageNet around task 25 which shows that this method cannot accumulate knowledge correctly past this point. Similarly, we can observe that ER also reaches a plateau around task 65 on Tiny ImageNet while no such behavior is displayed on CIFAR100. Such observation suggests that in this case, ER reaches maximum learning capabilities on Tiny ImageNet whereas not on CIFAR100. Interestingly, finetune's RAA is almost constant and near zero, which can be expected as such a naive approach shows no plasticity capabilities.

\begin{figure}[t]
    \centering
    \includegraphics[width=1.0\linewidth]{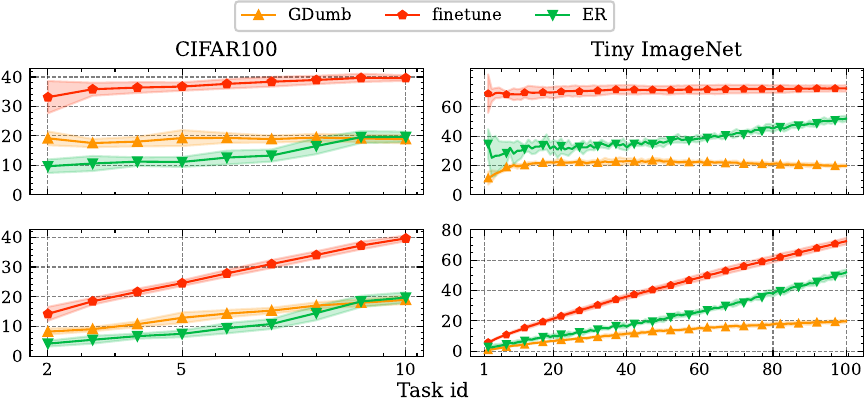}
    \caption{Values of $AF_k$ and $RAF_k$ with $k \in [1, K]$ the task id. From left to write, the performances are shown on CIFAR100 $M=2k$ with $K=5$ and Tiny ImageNet $M=2k$ with $K=5$. The top row shows the AF while the bottom row shows the RAF. Comparing AF to RAF, we can observe that for most cases $AF_k$ is rather constant while $RAF_k$ keeps increasing for every method. Best viewed in color.}
    \label{fig:af_all}
\end{figure}

\textbf{Increasing forgetting with RAF.} Figure~\ref{fig:af_all} shows AF and RAF values for considered methods. On both datasets, AF is rather constant or slightly increasing for all methods. However, RAF displays a strong increase throughout training for finetune and ER. Such an effect is less pronounced for GDumb. Even if no specific training regime can directly be observed with RAF, its overall trend highlights the different difficulty gaps between tasks. Indeed, a $60\%$ loss in accuracy from $\mathcal{T}_1$ to $\mathcal{T}_2$ should not be interpreted equally as a $60\%$ loss from $\mathcal{T}_{9}$ to $\mathcal{T}_{10}$. If we consider $10$ classes per task, the former corresponds to going from $10$ to $20$ classes to classify while the latter corresponds to going from $90$ to $100$ classes to classify. The difficulty gap is therefore much stronger in the former case than in the latter and should not translate as the same forgetting value.

\section{Conclusions}
\label{sec:conclusion}

In this paper, we discussed the limitations of current traditional CL metrics. After showing that AF and AA suffer from a setup-induced bias which can deteriorate continual learners analysis, we introduced new metrics for CL taking into account the increasing difficulty of the setup. We conducted several experiments on classical methods for the Online Class Incremental Learning scenario and experimentally demonstrate various advantages of using RAA and RAF for analyzing continual learners. We believe these metrics can help design new methods for CL. Extensions of this work might consider expanding the notion of task difficulty beyond the performance of a random classifier.




\bibliographystyle{miru2023e}
\bibliography{main}


\clearpage

\appendix
\section{Proof of proposition 4.3}
\label{appen:proof_prop_raa}

\begin{lemma}
\label{lem:aa_rand} $AA_k(Rand_{C_k})=\frac{1}{C_k}$
\end{lemma}
\textit{Proof}. Trivial.

\begin{lemma} $\max\limits_{f \in \mathcal{G},k}{uRAA_k(f)}=C_K$. With $K$ the total number of tasks.
\end{lemma}
\textit{Proof.} From definition 4.1 we have:
\begin{gather*}
    \max\limits_{f \in \mathcal{G},k}{uRAA_k(f)}=\max\limits_{f \in \mathcal{G},k}\frac{AA_k(f)}{AA_k(Rand_{C_k})}
\end{gather*}
With definition 3.1 and lemma \ref{lem:aa_rand}:
\begin{gather*}
    \max\limits_{f \in \mathcal{G},k}\frac{AA_k(f)}{AA_k(Rand_{C_k})}=\max\limits_{f \in \mathcal{G},k}\frac{C_k}{k}\sum_{j=1}^k{a_{k,j}(f)}
\end{gather*}
$\forall f \in \mathcal{G}$, $a_{k,j}(f) \in \lbrack0,1\rbrack$, so for any $k$, $uRAA_k(f)$ is maximum when $\forall (k,j) \in \lbrack1,K\rbrack^2$, $a_{k,j}(f)=1$. This corresponds to having a perfect classifier. Then we have:
\begin{equation}
\label{eq:max_uraa}
    \max\limits_{f \in \mathcal{G},k}{uRAA_k(f)}=\max\limits_{k}C_k= C_K
\end{equation}
Because the number of classes to classify increases linearly over time in our setup.\\
With definition 4.1 and equation \ref{eq:max_uraa}, we obtained the desired result.

\section{Proof of proposition 4.4}
\label{appen:proof_prop_raf}
\begin{lemma}
\label{lem:af_rand}
$AF_k(Rand_{C_k})=\frac{1}{C_k}(\frac{k}{k-1}H_{k-1} - 1)$. With $H_k=\sum_{i=1}^k\frac{1}{i}$ the k-th harmonic number.
\end{lemma}
\textit{Proof}. From definition 3.2:
\begin{gather*}
\begin{aligned}
    AF_k(Rand_{C_k})&=\frac{1}{k-1}\sum_{j=1}^{k-1}f_{k,j}(Rand_{C_k})\\
    \text{with} \ \ \
    f_{k,j}(Rand_{C_k})&=\max\limits_{l\in\{j,\cdots,k-1\}}{(a_{l,j}(Rand_{C_l}))} - a_{k,j}(Rand_{C_k}) \\
                   &=\max\limits_{l\in\{j,\cdots,k-1\}}{\left(\frac{1}{C_l}\right)} - \frac{1}{C_k} \\
                   &=\frac{1}{C_j} \frac{1}{C_k} \\
                   &=\frac{C_k - C_j}{C_jC_k} \\
   f_{k,j}(Rand_{C_k})&=\frac{C_{k-j}}{C_jC_k}
\end{aligned}
\end{gather*}
Because $C_{k+1}=C_k+n_C$ with $n_C$ the number of classes per task.\\
Then we have:
\begin{gather*}
\begin{aligned}
    AF_k(Rand{C_k})&=\frac{1}{k-1}\sum_{j=1}^{k-1}\frac{C_{k-j}}{C_jC_k} \\
                    &=\frac{1}{(k-1)C_k}\sum_{j=1}^{k-1}\frac{k-j}{j} \\
                    &=\frac{k}{(k-1)C_k}\sum_{j=1}^{k-1}\frac{1}{j} - \frac{1}{C_k} \\
                    &= \frac{1}{C_k} \left( \frac{kH_{k-1}}{k-1} -1 \right)
\end{aligned}
\end{gather*}
With $H_k=\sum_{j=1}^{k}\frac{1}{j}$. Falling back to definition 4.1:
\begin{gather*}
\begin{aligned}
    uRAF_k(Rand{C_k})&=\frac{AF_k(g)}{AF_k(Rand{C_k})} \\
                        &= \frac{C_k}{\frac{k}{k-1}H_{k-1}-1}\frac{1}{k-1}\sum_{j=1}^{k-1}f_{k,j}(g) \\
                        &=\frac{C_k}{k(H_{k-1}-1)+1}\sum_{j=1}^{k-1}f_{k,j}(g) \\
                    &=\frac{C_k}{k(H_{k}-1)}\sum_{j=1}^{k-1}f_{k,j}(g)
\end{aligned}
\end{gather*}
\begin{lemma}
\label{lem:max_raf}
    $\max\limits_{h \in \mathcal{G},k}{uRAF_k(h)}=\frac{C_k(K-1)}{K(H_{K}-1)}$. With $K$ the total number of tasks.
\end{lemma}
\textit{Proof.}
$\forall h \in \mathcal{G}$, $f_{k,j}(h) \in \lbrack0,1\rbrack$, so for any $k$, $uRAF_k(h)$ is maximum when $\forall (k,j) \in \lbrack1,K\rbrack^2$, $f_{k,j}(h)=1$. This corresponds to having the worst classifier. Then we have:
\begin{gather*}
    \max\limits_{h \in \mathcal{G},k}{uRAF_k(h)}=\max\limits_{k}\frac{C_k(k-1)}{k(H_k-1)}=\frac{C_K(K-1)}{K(H_K-1)}
\end{gather*}
Because $C_k=kn_C$ with $n_C$ the number of classes per task and $\frac{k-1}{H_k - 1}$ is an increasing function. \\
Eventually with \ref{lem:max_raf} we have:
\begin{gather*}
\begin{aligned}
    RAF_k(g)&=\frac{uRAF_k(g)}{\max\limits_{h \in \mathcal{G},k}{uRAF_k(h)}}=\frac{KC_k(H_K-1)}{kC_K(H_k-1)(K-1)}\sum_{j=1}^{k-1}f_{k,j}(g) \\
            &=\frac{H_K-1}{(H_k-1)(K-1)}\sum_{j=1}^{k-1}f_{k,j}(g)
\end{aligned}
\end{gather*}

\end{document}